%% file: paper.tex
\documentclass[10pt,twocolumn,letterpaper]{article}

\usepackage{cvpr}
\usepackage{times}
\usepackage{epsfig}
\usepackage{graphicx}
\usepackage{amsmath}
\usepackage{amssymb}
\usepackage{color}
\usepackage{ulem}
\usepackage{url}

\newcommand{\mysubsubsubsection}[1]{\vspace{0.1cm} \noindent {\bf #1}:}

\ifx\pdfoptionalwaysusepdfpagebox\relax\else
\fi

\usepackage[pagebackref=true,breaklinks=true,letterpaper=true,colorlinks,bookmarks=false]{hyperref}

\cvprfinalcopy 

\def\cvprPaperID{0} 

\ifcvprfinal\fi
\begin{document}

\title{Panoramic Structure from Motion via Geometric Relationship Detection}

\author{
Satoshi Ikehata  and  Yasutaka Furukawa\\
Washington University in St. Louis\\
\and
Ivaylo  Boyadzhiev
 and Qi Shan\\
Zillow\\
}

\maketitle

\input{abstract}

\input{introduction}

\input{related_work}

\input{input_data}
\input{sfm}
\input{sfm1}
\input{sfm2}
\input{sfm3}

\input{experimental_results}
\input{conclusions}

%

\vspace{0.3cm}
\noindent
{\large {\bf Acknowledgement}}

\noindent This research is partially supported by National Science
Foundation under grant IIS 1540012 and Zillow gift money. We also thank
Nvidia for a generous GPU donation. Yasutaka Furukawa is a consultant
for Zillow.

{\small
\bibliographystyle{ieee}
\bibliography{eccv16bib}
}
\end{document}


\title{Panoramic Structure from Motion via Geometric Relationship
Classification\\Supplementary material}

\author{
Satoshi Ikehata and Yasutaka Furukawa\\
Washington University in St. Louis\\
\and
Ivaylo  Boyadzhiev
 and  Qi Shan\\
Zillow\\
}

\maketitle

\input{supplement_algorithm}

\bibliographystyle{ieee}
\bibliography{eccv16bib}

%% file: abstract.tex
\begin{abstract}
This paper addresses the problem of Structure from Motion (SfM) for
indoor panoramic image streams, extremely challenging even for the
state-of-the-art due to the lack of textures and minimal parallax.
The key idea is the fusion of single-view and multi-view reconstruction
techniques via geometric relationship detection (e.g.,
detecting 2D lines as coplanar in 3D). Rough geometry suffices to
perform such detection, and our approach utilizes rough surface
normal estimates from an image-to-normal deep network to discover
geometric relationships among lines.
The detected relationships provide exact geometric constraints in our
line-based linear SfM formulation. A constrained linear least squares is
used to reconstruct a 3D model and camera motions, followed by the
bundle adjustment.
We have validated our algorithm on challenging datasets, outperforming
various state-of-the-art reconstruction techniques.
%
\end{abstract}

%% file: introduction.tex
 \section{Introduction}


Panorama images are everywhere on the Internet, instantly taking you to
remote locations such as Rome, the Louvre, or Great Barrier Reef under
water with immersive visualization.
%
%
%
%
Panoramas have become the first-class visual contents in digital
mapping, and are becoming increasingly more important with the emergence
of Virtual Reality.
%
%
%
%
%

\begin{figure}[tb]
	\begin{center}
		\includegraphics[width=\columnwidth]{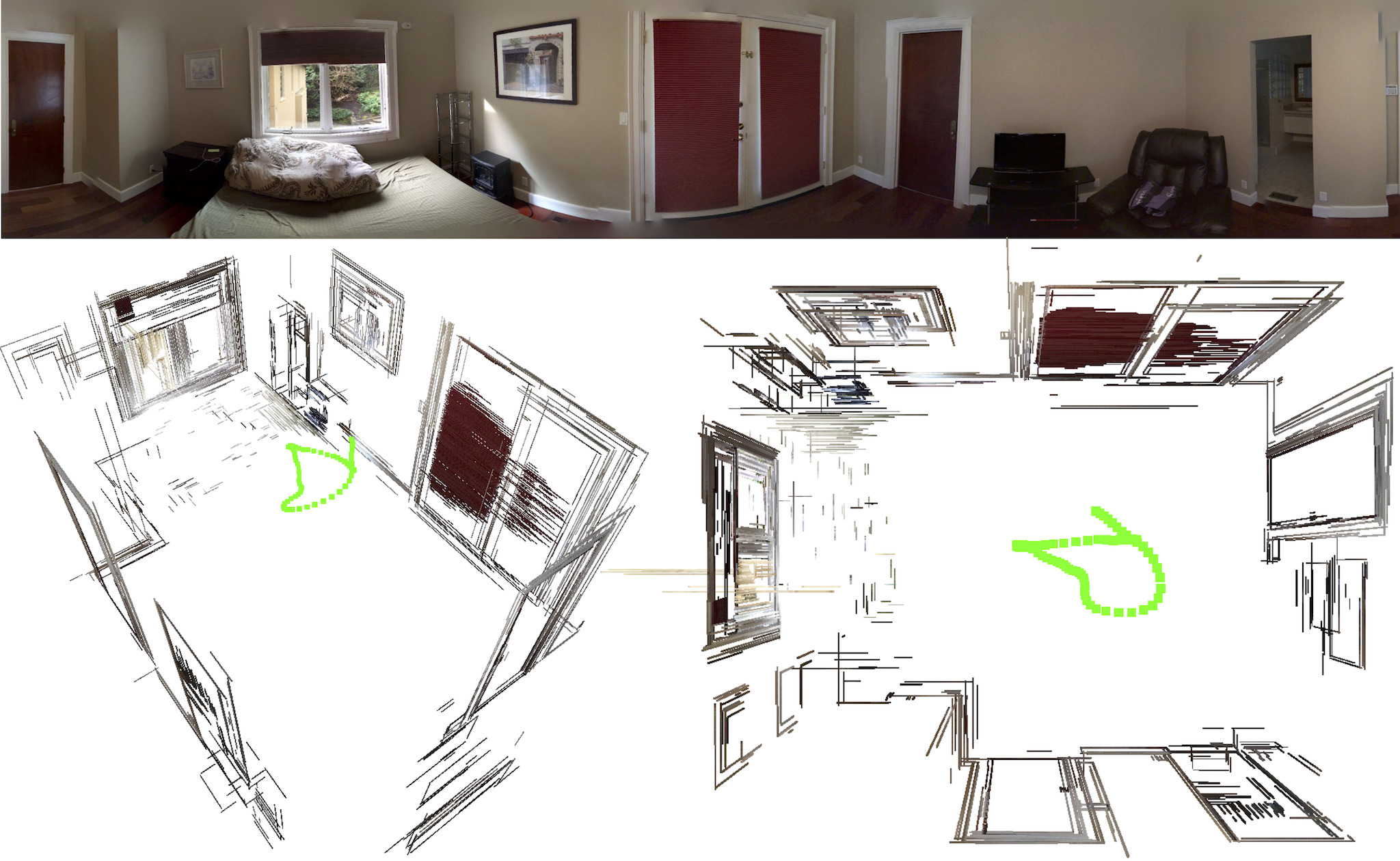}
	\end{center}
 \vspace{-0.2cm}
	\caption{Top: High quality panorama generation requires minimal
 camera translations, which make multi-view reconstruction
 difficult. Bottom: Our algorithm fuses single-view and multi-view
 reconstruction techniques to solve challenging SfM problems.
%
 } \label{fig:teaser}
\end{figure}

Panoramas, if equipped with the depth information, could enable 1) full
stereoscopic VR experiences; 2) 3D modeling of surrounding environments;
and 3) better scene understanding.
%
%
%
However, panoramic 3D reconstruction has been a challenge for Computer
Vision due to the minimal parallax, which is important to reduce
stitching artifacts~\cite{szeliski2010vision} but makes it difficult to
utilize powerful multi-view reconstruction techniques. The lack of
texture exacerbates the situations for indoor scenes.
The reconstruction accuracy of single-view methods is still far below
the production level
\cite{panocontext2014,merl2013junction,deep_depth,deep_normal}, and
successful panoramic 3D reconstruction has been demonstrated only with
the use of special hardware such as a depth
camera (e.g., Matterport), a camera array (e.g., Google Jump), a
spherical lightfield camera (e.g., Lytro Immerge), or a motorized tripod
constraining the motions~\cite{richardt2013megastereo}.

\begin{figure*}[tb]
 \includegraphics[width=\textwidth]{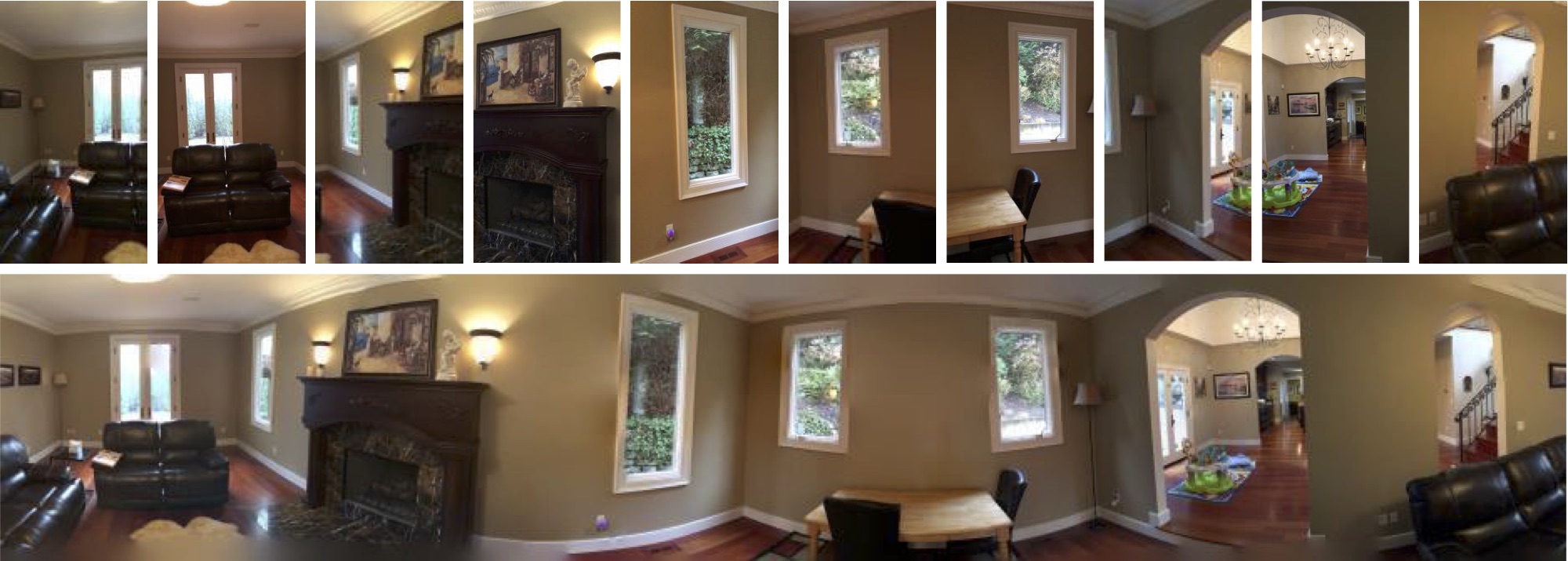}
 \caption{
 Top: Subsampled input frames. While they may not look particularly
 difficult, the challenge lies in the minimal baseline, where standard
 SfM or SLAM algorithms fail.
 Bottom: A stitched panorama image with minimal artifacts due to the
 small baseline.
 %
}  \label{fig:input_data}
\end{figure*}

This paper proposes a novel Structure from Motion (SfM) algorithm for
indoor panoramic image streams acquired by standard smart phones or
tablets (See Fig.~\ref{fig:teaser}). The key idea is the fusion of
single-view and multi-view reconstruction techniques.
In the past, 3D vision community has rarely seen such fusion,
mainly because single view methods are too ``rough'' to be
directly used with the multi-view techniques
(with some exceptions~\cite{xiao2013sun3d}).
We seek to utilize single-view techniques to effectively detect
geometric relationships of lines (e.g., detecting 2D lines as coplanar
in 3D),
which in turn yield precise geometric constraints to be used in
multi-view 3D reconstruction.
%
%
%
For example, once we identify floor image regions, we can declare
coplanarity among all the points or lines inside the floor regions,
providing powerful geometric constraints in solving for structure and
camera motions.
We formulate an SfM algorithm that expresses these geometric constraints
as linear functions of our variables, and uses a constrained linear
least squares to reconstruct a 3D model and camera motions, followed by
the bundle adjustment.
%
%
%
%
%
We have evaluated the proposed approach on many challenging datasets.
The qualitative and quantitative evaluations have demonstrated the
advantages of our method over many state-of-the-art reconstruction techniques.

%% file: related_work.tex
\section{Related work}

This paper proposes a novel SfM algorithm that integrates single-view
and multi-view reconstruction techniques, while utilizing geometric
relationships as indoor structure priors.
%
Therefore, we describe existing work on the following three domains:
1) single-view techniques for 3D reconstruction, 2) multi-view
reconstruction techniques, and 3) the use of structure priors for 3D
reconstruction.


\mysubsubsubsection{Single view technique}
A careful analysis of lines, its connections, and its vanishing points
has enabled a single-image reconstruction of architectural
scenes~\cite{merl2013junction,ramalingam2013lifting}.
However, these algorithms critically depend on the connectivity of lines
and can easily fail.
For indoor scenes, a single image scene understanding has been a very
active topic~\cite{hedau2009recovering,SilbermanECCV12,panocontext2014}.
%
%
Similar work exists for outdoor scenes~\cite{gupta2010blocks}. However,
their 3D models are mostly a combination of boxes for the purpose of
scene understanding rather than reconstruction. Superpixel and line
analysis allows more complicated reconstruction of an indoor
scene~\cite{yang2016efficient}.
More recently, data driven approaches, in particular deep networks, have
demonstrated interesting single-view reconstruction
results~\cite{deep_normal,mallya2015learning}.
Despite being an exciting new direction, these reconstructions are rough
and do not match up with the quality of multi-view reconstructions.

\mysubsubsubsection{Multi-view technique}
%
%
%
State-of-the-art SfM algorithms work very well for
texture-rich scenes with reasonable baselines. In 2006, Snavely et
al. introduced a powerful ``Incremental SfM'' algorithm~\cite{bundler},
which incrementally grows SfM models.
In 2011, Crandall et al. proposed a global approach, which seeks to
estimate all the camera parameters
simultaneously~\cite{crandall2011discrete}.
Currently, many state-of-the-art SfM or SLAM (Simultaneous Localization
and Mapping) algorithms follow this ``global approach''.


Shum and Szeliski in the nineties~\cite{Shum99} or Richardt et al. more
recently~\cite{richardt2013megastereo} have demonstrated
panoramic 3D reconstruction from rotation-dominant motions.
%
Yu et al.~\cite{dave2014smallbaseline}
or Ha et al.~\cite{ha2016high} succeeded in solving SfM from accidental
camera motions.
%
However, these methods assume rich texture and rely on visual
feature tracking/matching with careful multi-view geometric analysis for
3D reconstruction.~\footnote{Google Cardboard
Camera App produces stereoscopic panoramas from
panoramic movies. Although its algorithmic details are not
disclosed,
feature tracking/matching is probably their primary geometric cues.}
Texture-less scenes with limited camera translations still pose major
challenges for existing SfM algorithms.

\mysubsubsubsection{Structure priors for 3D reconstruction}
Structure priors have played an important role in the
advancement of 3D Computer Vision.
%
%
Flint et al.~\cite{flint2010dynamic} has proposed an effective indoor
scene reconstruction algorithm by assuming that an indoor scene consists
of two horizontal surfaces and vertical walls. This type
of high-level structural priors have also been the key to the success of
single-view reconstruction techniques.
However, in the domain of SfM or SLAM, only low-level geometric priors
have been exploited in the past, often just lines and vanishing
points~\cite{taylor1995structure,jeong2006visual,schindler2006line,elqursh2011line,kim2014planar}
or planes at best~\cite{salas2014dense}.
%
%
In this paper, we seek to exploit a family of geometric relationships
between lines to better constrain challenging reconstruction
problems.

%% file: input_data.tex
\section{Input data} \label{section:input_data}

\begin{figure*}[tb]
 \centerline{
 \includegraphics[width=0.80\textwidth]{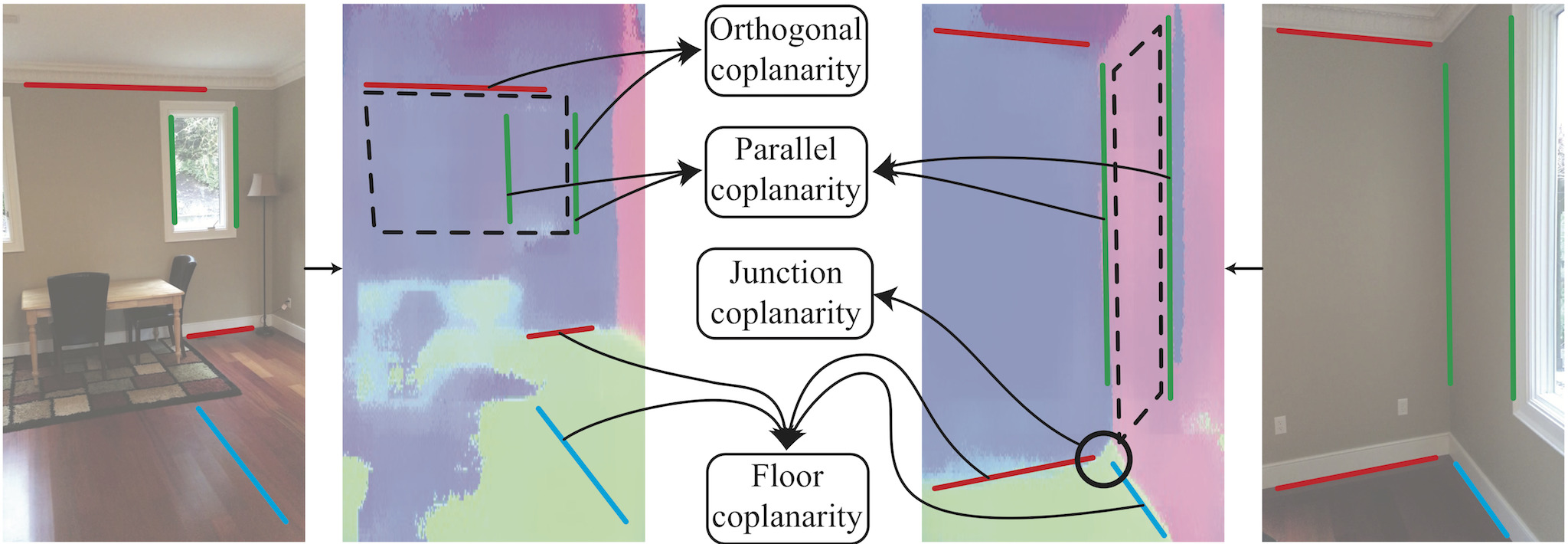} }
 \caption{We identify four different types of coplanarity relationships
 among line segments by utilizing a surface normal estimate by a
 deep-network~\cite{deep_normal}.
 The colors of line segments represent their corresponding Manhattan
 directions.}
 \label{fig:coplanar}
\end{figure*}

We have developed simple iOS and Android apps that record $360^\circ$
panoramic videos and IMU rotations (See Fig.~\ref{fig:input_data}). Both
iOS and Android offer an API to retrieve camera rotations after sensor
fusion.
%
Similar to Google Cardboard Camera App, we have recorded videos with a
``natural'' body motion in which the human body is at the center of
rotation, as opposed to an
``unnatural'' motion where the camera must be at the center of
rotation. This camera motion ensures some amount of parallax, although
being too small for standard SfM or SLAM algorithms.
Our approach identifies geometric constraints to enable 3D
reconstruction even from such small translational motions.


%


%% file: sfm.tex
\section{Panoramic Structure from Motion}

The input to our pipeline is a panoramic image stream with initial
camera rotations from the iOS or Andorid app as well as intrinsics from
precalibration~\cite{BouguetCalib}. The pipeline consists of four steps:
1) preprocessing, 2) geometric relationship detection, 3) linear SfM,
and 4) bundle adjustment. The second and third steps are the core of
this paper.
We now explain the details of each step.

\subsection{Preprocessing}
The goal of the preprocessing step is three-fold: Manhattan direction
extraction, Manhattan line tracking, and camera rotation refinement.
The procedure is based on standard techniques, and we here briefly
describe the procedure, and refer some algorithmic and implementation
details to the supplementary material.

\mysubsubsubsection{Line segment detection} First, we use a standard
line segment detection software (LSD~\cite{von2008lsd}) to extract line
segments from each input image. We use the existing
algorithm~\cite{Tavares1995} to merge neighboring line segments when
their angle differences are less than 1 degree and the minimum distance
between their endpoints is less than $0.05\times min(w,h)$ pixels where
$(w, h)$ are the width and height of the input image,
respectively. After the merging, we discard line segments that are
shorter than $0.05\times min(w,h)$ pixels.

\mysubsubsubsection{Manhattan frame extraction} Given reasonable initial
camera rotations and intrinsics, we extract the Manhattan frame from the
detected lines in all the images: 1) We let each line cast votes to its
potential 3D directions (i.e., a great circle) on a Gaussian sphere;
then 2) sequentially extract Manhattan directions by detecting peaks
while enforcing the mutual orthogonality. 

\mysubsubsubsection{Manhattan line extraction} We detect {\it Manhattan
line segments} by simply collecting lines that cast votes to each of the
three peaks within a certain margin (10 degrees on the voting sphere).
We avoid detecting degenerate lines that are associated with two or more
Manhattan directions.

\mysubsubsubsection{Rotation refinement} Since initial rotations from IMU usually contain drifting errors, we use standard non-linear least
squares optimization~\cite{ceres} to refine camera rotations so that the
detected Manhattan line segments pass through the corresponding
Manhattan vanishing points.
We repeat the Manhattan axis extraction, Manhattan line extraction, and
rotation refinement a few times.

\mysubsubsubsection{Line tracking} Lastly, we form tracks of Manhattan
line segments by grouping nearby line segments along the same Manhattan
direction across frames. 
%
%
%



%% file: sfm1.tex
\subsection{Geometric relationship detection}\label{sec:coplanarity}

Rough surface normal estimations suffice to extract powerful geometric
constraints among lines. We first use a deep-network by Wang et
al.~\cite{deep_normal} to obtain the surface normal estimation for each
input image. Since estimated surface normals are defined on each camera
coordinate frame, we project each normal onto the global Manhattan
coordinate system using the rotation matrices.

Three types of coplanarity relationships are detected for every pair of
line segments in each frame. The fourth coplanarity test finds
and enforces line segments on the floor to be coplanar, providing
precise geometric constraints across all the frames even without any
visual overlap
(See Fig.~\ref{fig:coplanar}).

\begin{figure}[tb]
 \centerline{
 \includegraphics[width=0.80\columnwidth]{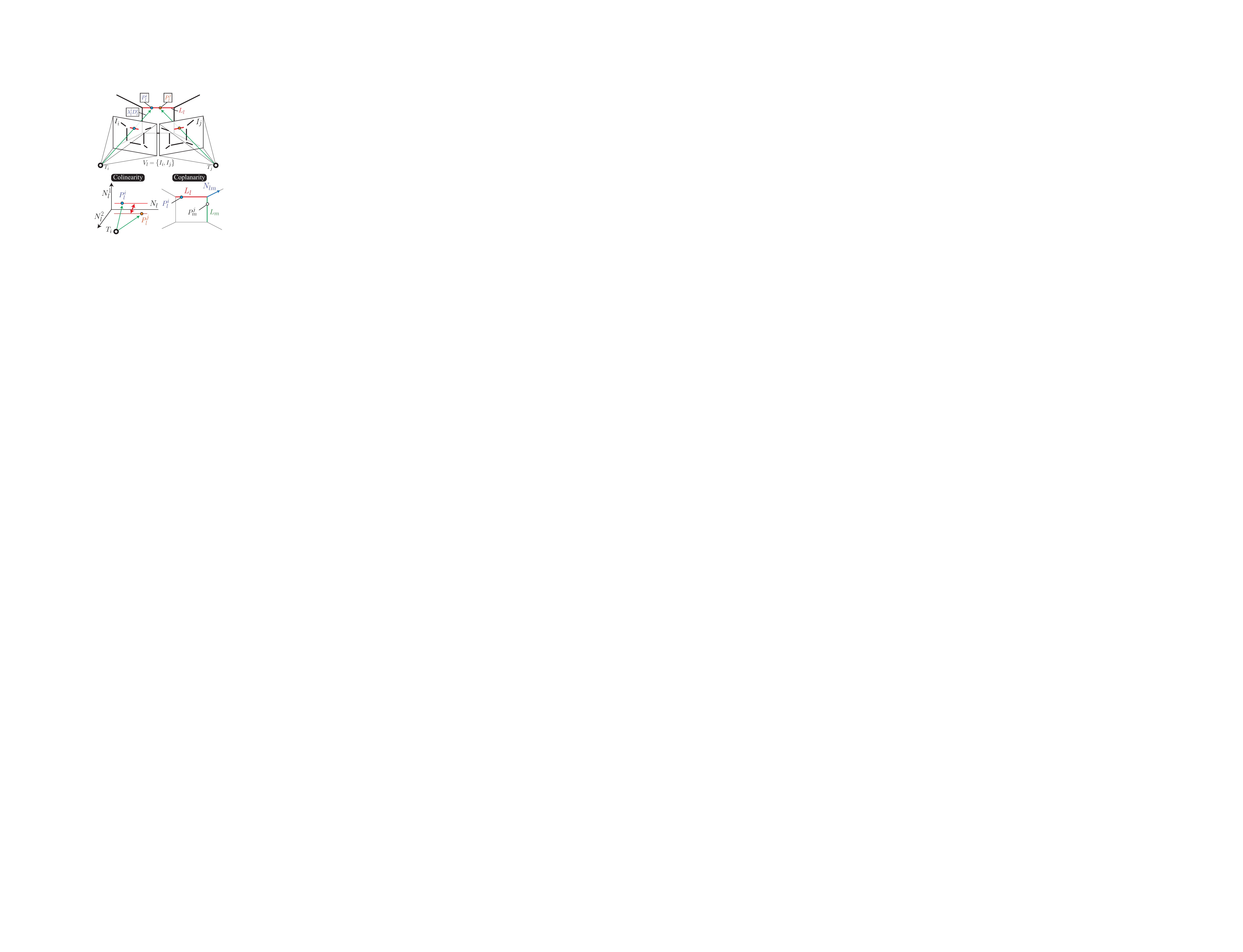} }
 \caption{Linear SfM formulation with geometric constraints.  Suppose a
 line $L_l$ was detected and formed a track in images $V_l=\{I_i,
 I_j\}$. The mid-point ($P^i_l$) of a line segment in an image ($I_i$)
 is parameterized by the depth ($\lambda^i_l$). The line direction is a
 known Manhattan direction. We enforce detected coplanarities between a
 pair of lines as well as the colinearity between $P^i_l$ and $P^j_l$.
 } \label{fig:notation}
\end{figure}


\begin{figure*}[!t]
         \includegraphics[width=\textwidth]{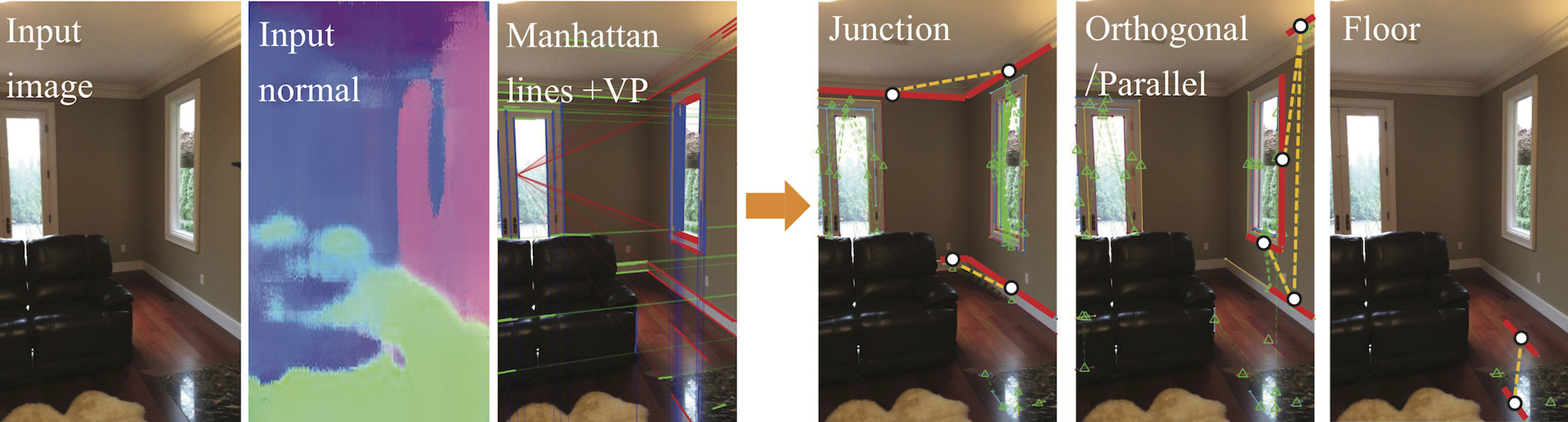}
 \caption{Geometric relationship detection. Given a video sequence and a
 set of normal images (1st and 2nd images), our framework estimates the
 Manhattan-world vanishing points and cluster each line segment into one
 of three three Manhattan axes (3rd image). Then, we detect four
 different types of coplanarity relationships among line segments:
 junction, orthogonal, parallel, or floor coplanarity.
 To avoid clutter, we only show a small number of detected geometric
 relationships.
 }  \label{fig:intermediate2}
\end{figure*}
\mysubsubsubsection{Orthogonal coplanarity} Our input is Manhattan
lines, each of which is associated with one of the three Manhattan
directions.
Suppose one is given a pair of lines associated with different Manhattan
directions.  If the pair is coplanar, the space between these two lines
should have the same surface normal pointing towards the orthogonal
Manhattan direction. Therefore, we compute the average surface normal
inside a quad defined by the four end-points of the line segments, then
declare Manhattan coplanarity if the following two conditions are met:
1) the average normal is within $20^\circ$ from the expected Manhattan
direction and 2) the average angle difference between the average normal
and all the surface normals inside the quad is less than $5^\circ$.


\mysubsubsubsection{Parallel coplanarity}
Given a pair of lines with the same Manhattan direction, we
detect the coplanarity in exactly the same way as in the Orthogonal
coplanarity case with one difference. Parallel lines are always
coplanar, and we restrict the potential coplanarity normal to be one of
the remaining two Manhattan directions.
For instance, if two lines are parallel along the X-axis, the
coplanarity normal must be either along Y or Z axis.
%

%
%

\mysubsubsubsection{Junction coplanarity}
Given a pair of lines with different Manhattan directions, two
lines are deemed to be coplanar if their end-points are close, that is,
within $0.1\times min(h,w)$.


\mysubsubsubsection{Floor coplanarity} The floor coplanarity segments
the floor region in each frame by collecting pixels whose normals are
within $25^\circ$ degrees from the vertical
direction.  We apply a morphological operation (dilation) once then find
line segments that are fully contained inside the floor regions. We
enforce these lines from all the frames to be coplanar on a horizontal
surface (i.e., floor). Notice that existing SfM/SLAM algorithms require
features to be commonly visible across frames. The floor coplanarity is
unusually powerful and can provide constraints among all the frames even
when there are no visual overlap.







%% file: sfm2.tex
\subsection{Linear SfM formulation}

Sinha et al. proposed a linear SfM formulation that minimizes
reprojection errors as linear functions of the point 3D
coordinates and camera translations~\cite{SudiptaLinearSfM}.  We have
formulated a linear SfM problem that minimizes colinearity and
coplanarity constraints among lines as linear functions of the
line parameters and camera translations.

\mysubsubsubsection{Model} IMU rotations and the camera intrinsics from
the precalibration step allow us to focus on the estimation of camera
translations ($\{T_i\}$) and the 3D model, in our case, 3D lines. Since
we know the direction of a line (i.e., one of the Manhattan directions),
estimating the depth of a single reference point on a line suffices to
uniquely determine its geometry (See Fig.~\ref{fig:notation}). In
particular, we seek to estimate the depth $\lambda^i_l$ of a tracked
line at its mid-point $P^i_l$ in each image, where $i$ and $l$ are the
image and line indexes, respectively:
\begin{eqnarray*}
 P^i_l = T^i + \lambda^i_l D^i_l.
\end{eqnarray*}

$D^i_l$ is the unit-length viewing ray for $P^i_l$. The depths are
measured along the rays as opposed to along the optical axis of the
image.
Note that we estimate multiple depth values for a single line-track,
making our  line parameterization redundant.
However, we have chosen this simple parameterization because the core
solver (constrained linear least squares) is scalable.
In the bundle adjustment step conducting non-linear optimization next,
we will use more compact line parameterization.
Unknown variables of our SfM problem are camera translations and line
depths, subject to the following two linear constraints.

\mysubsubsubsection{Colinearity constraints} A single line-track has
multiple depth values estimated across tracked images. Lines must be
reconstructed exactly at the same location, and we enforce colinearity
among such lines. More precisely, give a line, for every pair of tracked
images ($I^i, I^j$), we measure the
distance between the two lines along their orthogonal Manhattan directions
($N^1_l, N^2_l$):
%
\begin{eqnarray*}
 \left(P^i_l - P^j_l\right) \cdot N^1_l = 0, \quad
 \left(P^i_l - P^j_l\right) \cdot N^2_l = 0.
\end{eqnarray*}

\mysubsubsubsection{Coplanarity constraints}
Let $P^i_l$ and $P^j_m$ be two lines that have been detected as
coplanar.
We impose coplanarity  as
\begin{eqnarray*}
 (P^i_l - P^j_m)\cdot N_{l,m} = 0.
\end{eqnarray*}
$N_{l,m}$ is the surface normal of the detected plane.

\vspace{0.2cm}
%

\noindent 
These constraints are linear functions of the variables ($\{T_i\}$,
$\{\lambda^i_l\}$.
As $\lambda$ (depth) must be positive, we add the non-negativity
constraint for the depth variables, then use the standard dual
active-set method in QPC~\cite{qpc}.
%




%% file: sfm3.tex
\subsection{Bundle adjustment}
Bundle adjustment further improves the quality of 3D models and camera
parameters.
This time, we employ more compact line parameterization introduced
in~\cite{taylor1995structure}. To be more specific, a line $L_l$ is
parameterized by a vector $\Lambda_l$ that connects the origin and the
closest point on the line. It is easy to show that $\Lambda_l$ becomes
perpendicular to the direction of a line. The residual again consists of
the three terms.

\mysubsubsubsection{Colinearity} Since $\Lambda_l$ is perpendicular to the
line direction, which we denote as $A_l$, their dot-product must be 0:
\begin{eqnarray*}
 A_l \cdot \Lambda_l = 0.
\end{eqnarray*}

\vspace{-0.2cm}
\mysubsubsubsection{Coplanarity} Given two lines that must be coplanar and
are parameterized by $\Lambda_l$ and $\Lambda_m$, respectively, the
distance of the two lines along the coplanar normal direction ($A_{lm}$)
must be 0:
\begin{eqnarray*}
 (\Lambda_l - \Lambda_m) \cdot A_{lm} = 0.
\end{eqnarray*}

\vspace{-0.2cm}
\mysubsubsubsection{Reprojection errors} We project a line $L_l$ to all
its tracked images and measure the reprojection errors against the line
segments in the images. There are many ways to measure the line
reprojection errors. We simply measure the average distance over all the
points on the line segment to the projected
line~\cite{taylor1995structure}.

\vspace{0.2cm}
We use a standard non-linear least squares optimization library
Ceres~\cite{ceres} to refine the parameters. The weights of the three
residuals are set to $1$ for the reprojection error and $10^4$ for the
colinearity and coplanarity errors.
Following recent trends in SfM literature~\cite{openMVG}, we solve
this bundle adjustment problem in three phases. In the first phase, we
only refine line parameters and camera translations. We then add camera
rotations in the second phase, and camera intrinsics in the last phase
for refinement.


%% file: experimental_results.tex
\section{Experimental results}


\begin{figure*}[tb]
 \includegraphics[width=\textwidth]{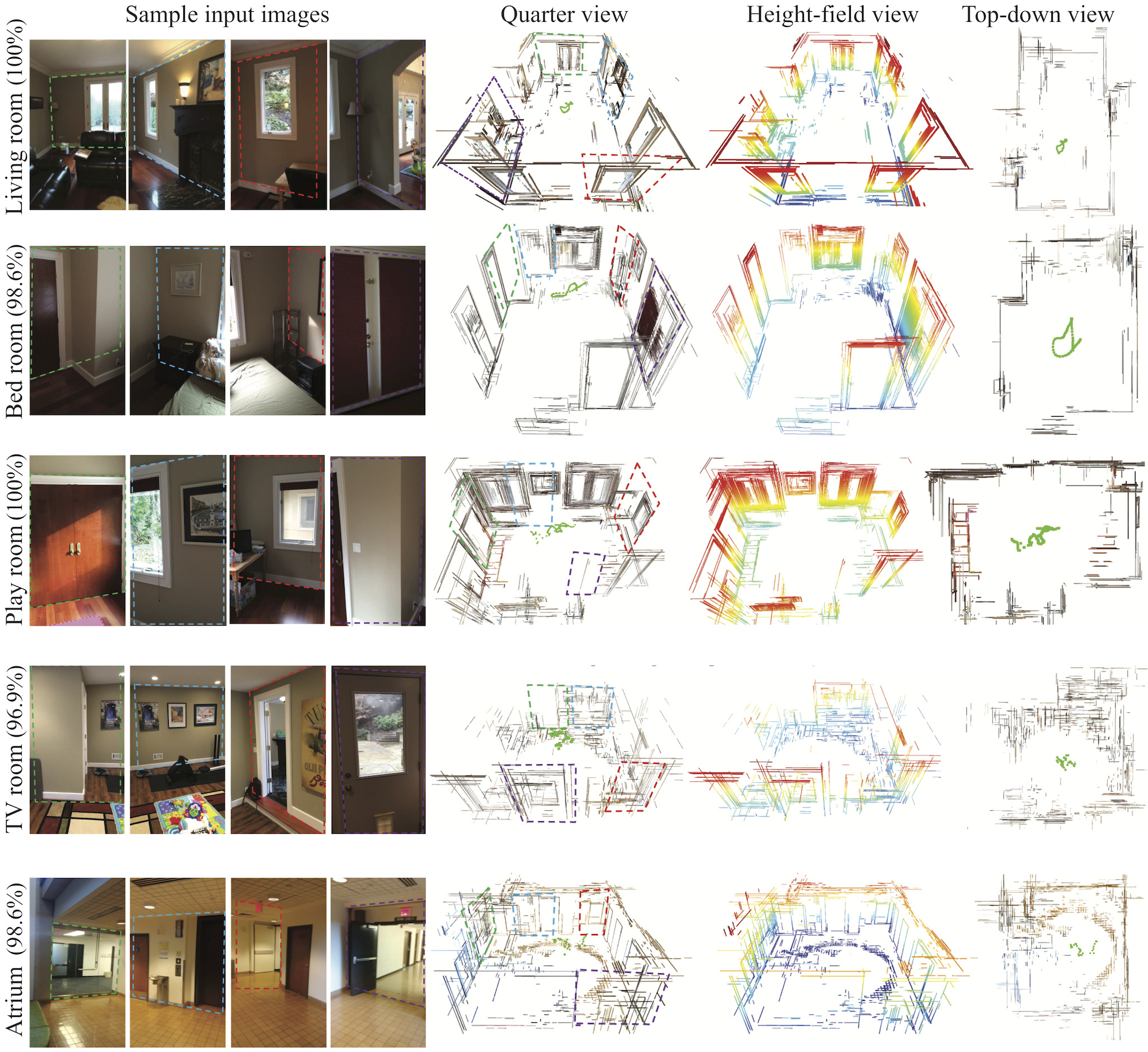}
 \caption{Sample input images and our line models in three different
 views.
 The quarter view renders each line with its average color in the
 images. The height-field view uses the heat-map color scheme based on
 the heights from the floor.
 The top-down view is an orthographic projection view from the
 top. Recovered camera centers are shown by green dots. The numbers show
 the ratios of registered frames.}  \label{fig:results2}
\end{figure*}

\begin{figure*}[tb]
 \includegraphics[width=\textwidth]{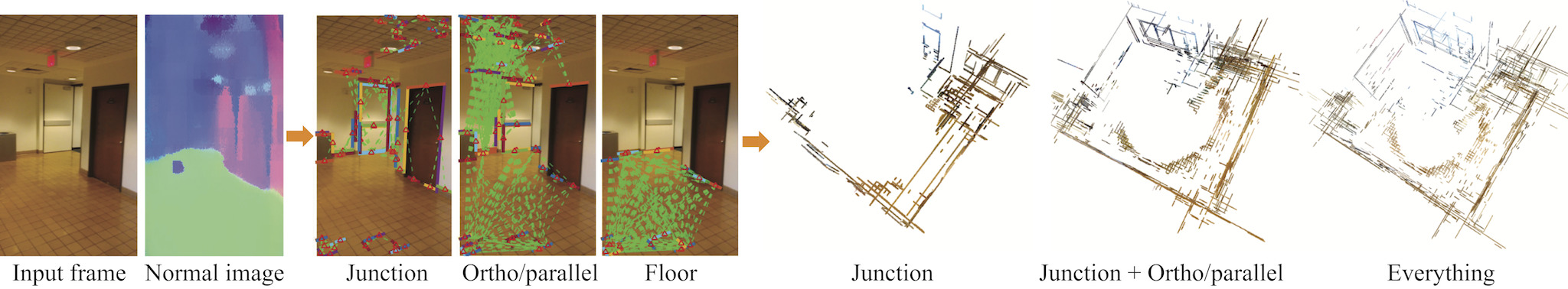}
 \caption{Contributions of the four coplanarity relationships, detected
 in the {\it Atrium} dataset.  We have run our reconstruction algorithm
 with three different sets of coplanarity relationships: 1) junction
 coplanarity only; 2) junction, orthogonal, and parallel coplanarities;
 and 3) all the four. The middle shows the pairs of line segments
 detected as coplanar.}  \label{fig:synth}
\end{figure*}


We have evaluated the proposed approach with five challenging
datasets.
%
The first four datasets have been captured in a residential house and
named {\it Living room} ($340$ frames), {\it Bed room} ($341$ frames),
{\it Play room} ($296$ frames), and {\it TV room} ($294$ frames). The
resolution of these datasets are $1980\times 1080$ (iPhone6s).
%
The fifth dataset was captured at an atrium of a University building:
{\it Atrium} ($361$ frames).
The resolution of this dataset is $1280\times 800$ (Nexus 9 tablet). All
the datasets have been recorded as 30 fps videos, and sub-sampled so
that $360^\circ$ are covered by roughly $360$ frames.

We have used a PC with an Intel Core i7-4770 ($3.40$GHz, single thread)
processor and $32.0$GB ram. MATLAB with some C++ mex functions have been
used for the implementation.  The rough processing times of our
computationally expensive steps are 1) 30 seconds for the line detection
and merging; 2) 1 minute for the vanishing point estimation and rotation
refinement; 3) 2 minutes for the line tracking; 4) 5 seconds per image
for the coplanarity estimation; 6) 30 seconds for solving a constrained
linear least squares problem; and 7) 1 minute for the bundle adjustment.


\begin{figure*}[tb]
 \includegraphics[width=\textwidth]{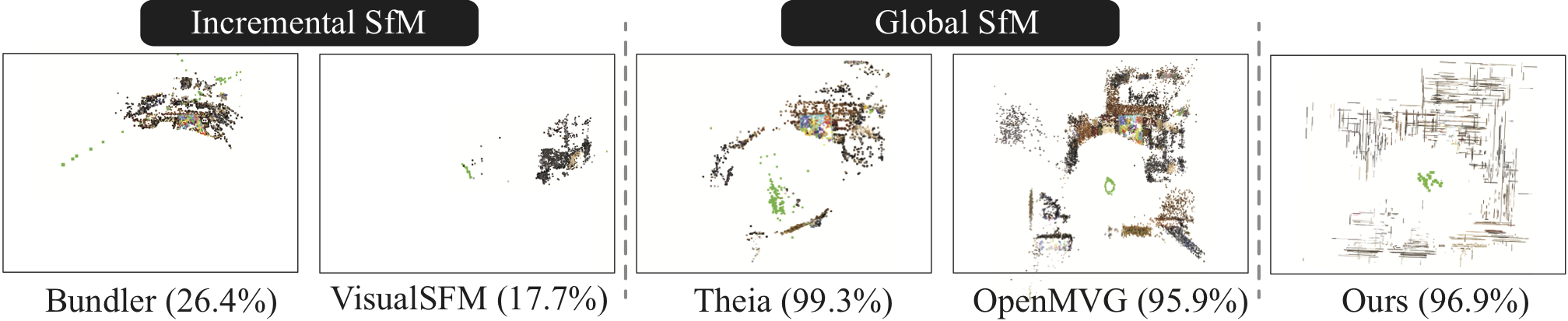}
 \caption{Comparison against state-of-the-art SfM (SLAM) systems,
 Bundler~\cite{bundler}, VisualSFM~\cite{VisualSFM},
 Theia~\cite{sweeney2015theia}, and OpenMVG~\cite{openMVG} for TV
 room. Each number shows the ratio of registered frames. Only OpenMVG
 and our approach have produced near perfect models.}
 \label{fig:bundler_visualsfm}
\end{figure*}

\begin{figure*}[tb]
\centerline{
 \includegraphics[width=\textwidth]{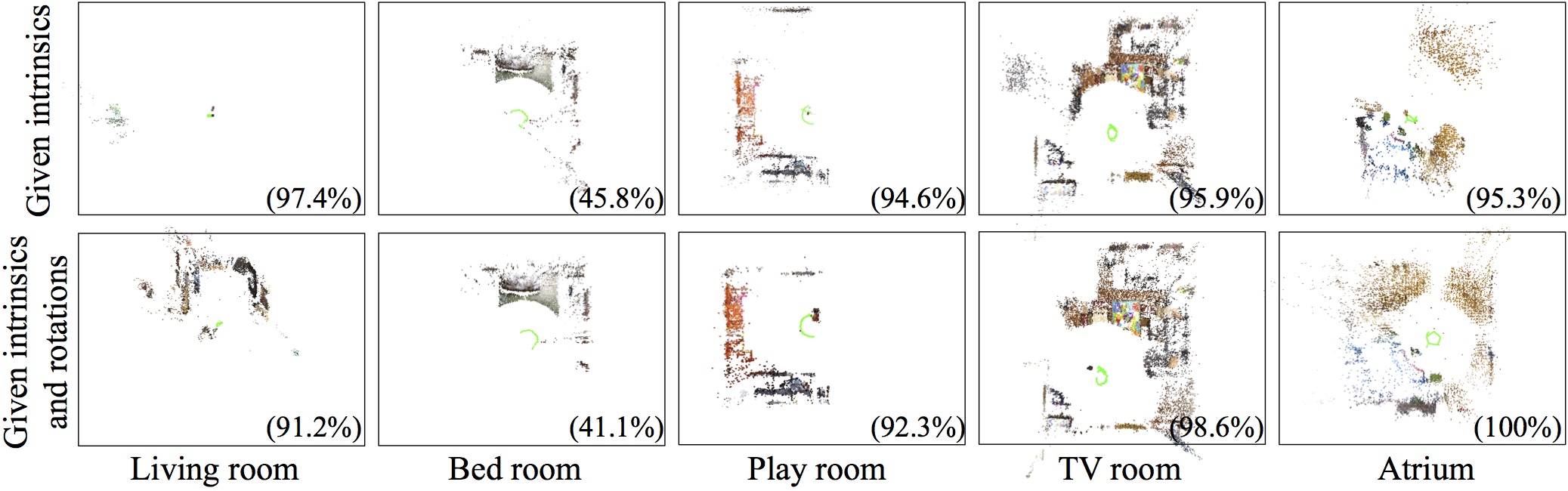} }
 \caption{OpenMVG results in a top-down view. For being fair, the
 intrinsics are initialized to the values in our pre-calibration process
 in the top row.
 Both the intrinsics and camera rotations are initialized in the
 bottom row. The numbers show the ratios of registered frames.} \label{fig:compOpenMVG}
\end{figure*}

The main experimental results are illustrated in
Fig.~\ref{fig:results2}.
%
%
%
%
The proposed approach has successfully recovered near-complete 3D
models, registering more than 90\% of the frames in each dataset.
%
Figure~\ref{fig:synth} illustrates the contributions of different
geometric relationships, where
we have run our algorithm on {\it Atrium} with three different sets of
relationships: 1) junction coplanarity only; 2) junction, orthogonal,
and parallel coplanarities; and 3) everything. In addition to making the
3D structure more accurate, the geometric relationships help connect
more images and models, which would otherwise be disconnected and become
scale-ambiguous, the major and most problematic failure modes of current
SfM methods.

We have compared against a wide range of SfM (SLAM) software to assess
challenges in our panoramic movie datasets. First,
Figure~\ref{fig:bundler_visualsfm} shows the reconstruction results of
the four state-of-the-art SfM/SLAM systems with ours for Play room,
which is a relatively easier dataset with rich textures. The left two
methods are so-called ``Incremental SfM'', which sequentially adds
cameras and grows the model. The next two methods are ``Global SfM'',
which simultaneously recovers all the camera parameters.
For fairness, camera intrinsics have been provided to each
software as either initialization or fixed parameters, except that we
could not figure out a way to specify in Theia.
As the figure illustrates, Global SfM is the state-of-the-art approach
and outperforms in this challenging example.  We have also tried to
evaluate small-motion SfM algorithms, in particular, DfUSMC by Ha et
al.~\cite{ha2016high}. However, they could not produce any models as the
software assumes that feature tracks must be fully visible throughout the
video. In rotation-heavy panoramic movies, features quickly go outside
frames and tracks become short, another challenge in our problem.

To further evaluate the effectiveness of our method,
Figure~\ref{fig:compOpenMVG} shows the reconstructed models of
OpenMVG~\cite{openMVG}, the best method in the previous experiment, on
all the datasets. In addition to the intrinsics (the top row), we have
also provided the IMU camera rotations as the initialization (the bottom
row). The ratios of registered frames suggest that OpenMVG has produced
near complete models for most of the examples. However, the effective
completeness of the OpenMVG models appear much lower, especially in the
left three examples, 
%
where scarce features and small translations pose challenges.

%
%

\begin{figure*}[tb]
 \centerline{ \includegraphics[width=\textwidth]{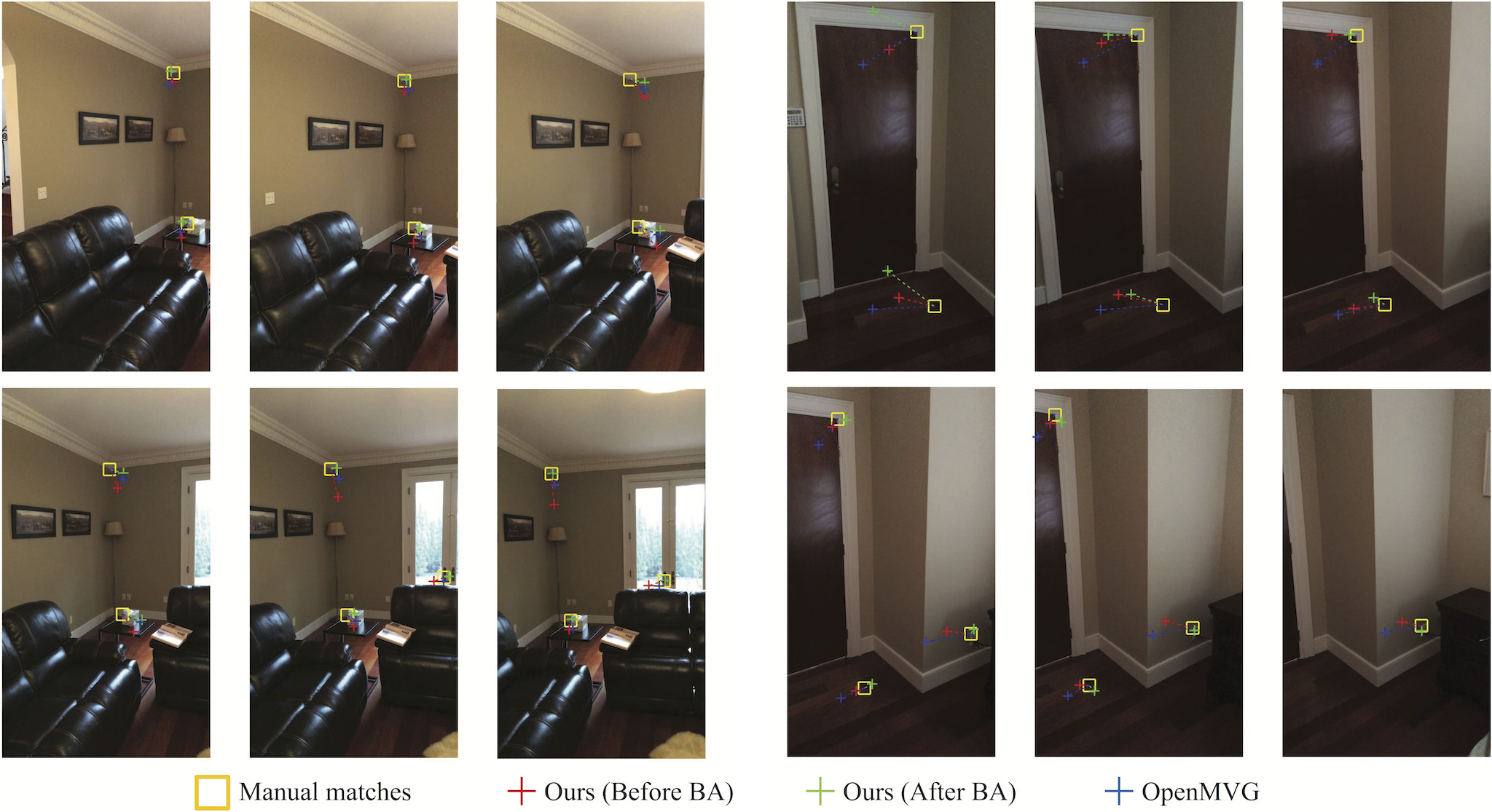} }
 \vspace{0.05cm}
 \centerline{
 \includegraphics[width=0.7\textwidth]{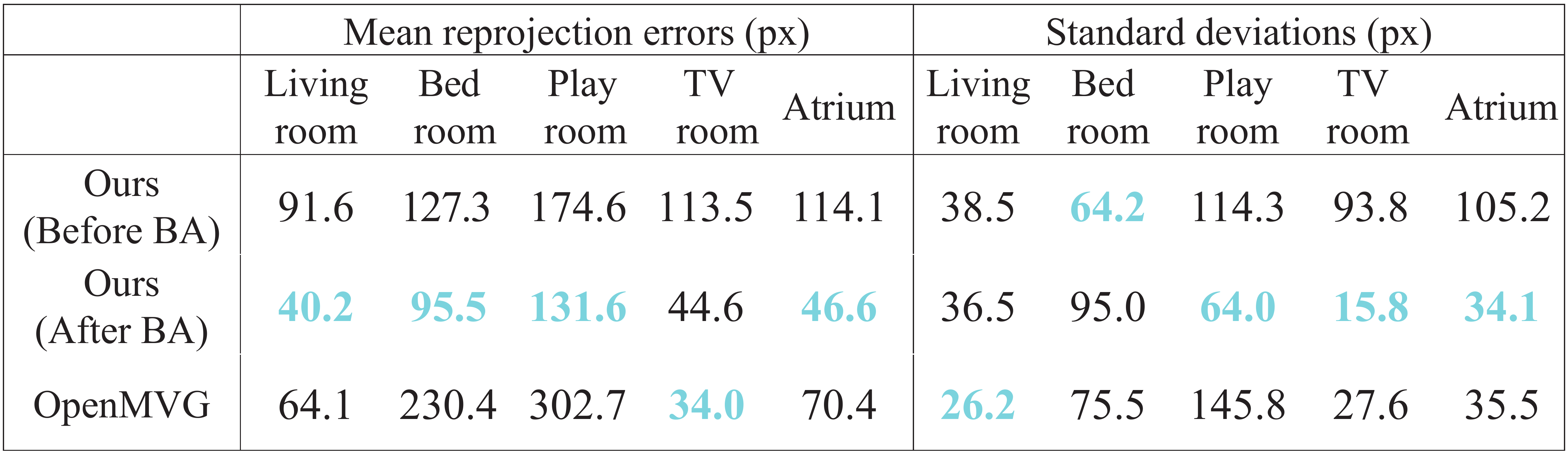} }
 \vspace{0.05cm}
 \caption{Reprojection error analysis in the {\it Living room} (left)
 and {\it Bed room} (right) datasets.  We have triangulated a 3D point
 from manual correspondences (yellow rectangles), then plot reprojected
 pixel coordinates based on the camera parameters of our method (before
 or after the bundle adjustment) and OpenMVG. The table shows the means
 and standard deviations of the reprojection errors in pixels.}
 \label{fig:quantitative}
\end{figure*}


\begin{figure}[tb]
 \begin{center}
  \includegraphics[width=\columnwidth]{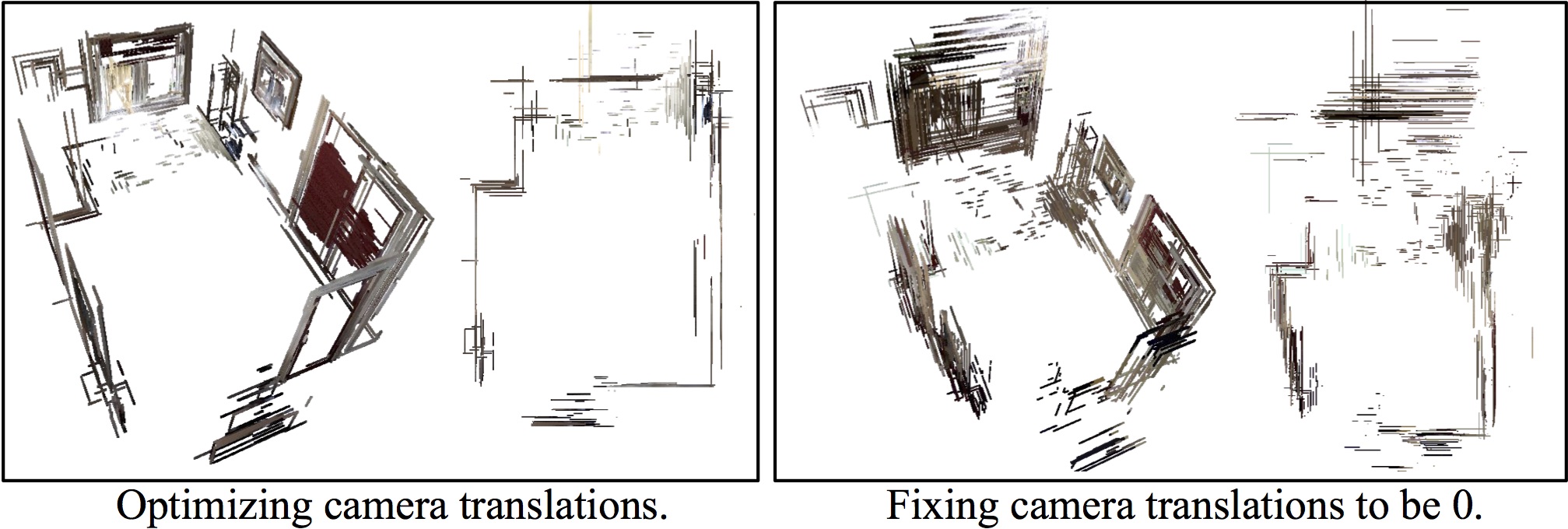}
 \end{center}
 \caption{Estimation of the camera translation in the SfM
 framework is crucial in obtaining a clean model.
 }  \label{fig:synth2}
\end{figure}

A quantitative evaluation of scene reconstructions with a ground-truth is a
challenging task especially for complex indoor environments. We have
manually clicked correspondences across multiple images, triangulated
the 3D point, and used its reprojection errors as the accuracy measure
(See Fig.~\ref{fig:quantitative}). More precisely, for each dataset, we
have selected six images with some visual overlap, where OpenMVG has
estimated camera parameters.  The table shows the means and standard
deviations of the reprojection errors in pixels. Note that the
resolution of the input images are $1980\times 1080$ except for the
Atrium dataset whose resolution is $1280\times 800$. Our indoor
panoramic image streams are extremely challenging as
distinctive visual feature points are often rare in a sequence, and
reprojection errors are relatively large throughout the sequences.
Nonetheless, our errors are often a few times smaller than those of
OpenMVG.



Our final experiment is to verify the importance of camera translation
estimation in the SfM framework. We have simply run our algorithm while
enforcing camera translations to be 0, which resembles a problem setting
for line-based single view reconstruction. Figure~\ref{fig:synth2} shows
that the translation estimation is crucial in obtaining a clean 3D model
without corruption. Please also see the supplementary video for the
full assessment of the input videos and the reconstruction results.
%

%
%




%% file: conclusions.tex
\section{Conclusions}

This paper tackles a challenging panoramic SfM problem, where input
images have minimal parallax and lack in rich visual textures. Our
approach detects coplanarity relationships between pairs of lines by
utilizing a deep-network for the surface normal estimation. The detected
relationships provide exact geometric constraints in solving a
line-based SfM problem. The presented method has outperformed many
state-of-the-art SfM (SLAM) algorithms on our challenging datasets.
%
%
%
%
%
The current limitation of our approach is the false coplanarity
detection. While 3D structure looks clean, reprojection errors are still
too large to run a stereo algorithm for obtaining a dense geometry. Our
future work is to 1) incorporate point features into the framework; 2)
train a proper relationship classification machinery given an image and
a pair of lines; and 3) develop robust optimization strategy to handle
outliers.
%



%% file: supplement_algorithm.tex
The supplementary material provides algorithmic and implementation
details of the three preprocessing steps: (1) Manhattan frame
(direction) extraction, (2) camera rotation refinement, and (3)
Manhattan line tracking. These steps rely on standard techniques and are
described here.

\section{Manhattan frame extraction}
Manhattan frame extraction aims to recover the three orthogonal
Manhattan directions given images, camera intrinsics and IMU rotation
matrices.

\begin{figure*}[tbh]
 \begin{center}
  \includegraphics[width=\textwidth]{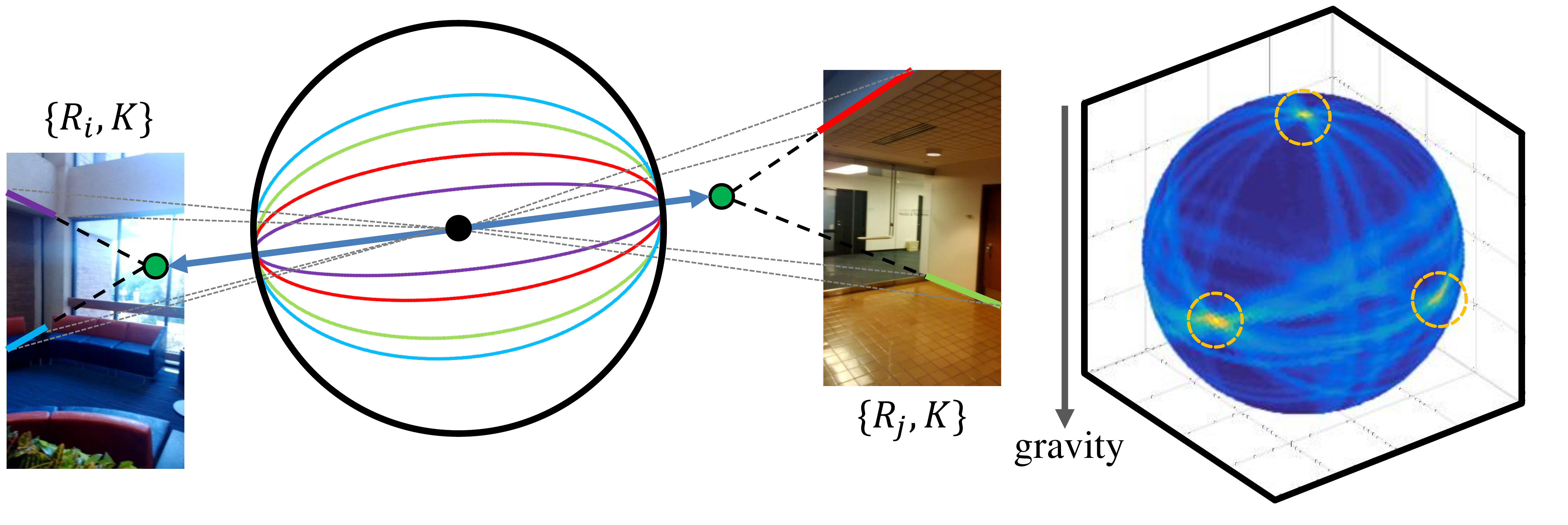}
 \end{center}
 \caption{Manhattan frame extraction. (left) We vote along the
 interpretation plane on the Gaussian sphere. (right) vanishing
 directions are extracted at the peaks on the sphere.}
 \label{fig:frameextraction}
\end{figure*}
\begin{figure*}[tbh]
 \begin{center}
  \includegraphics[width=\textwidth]{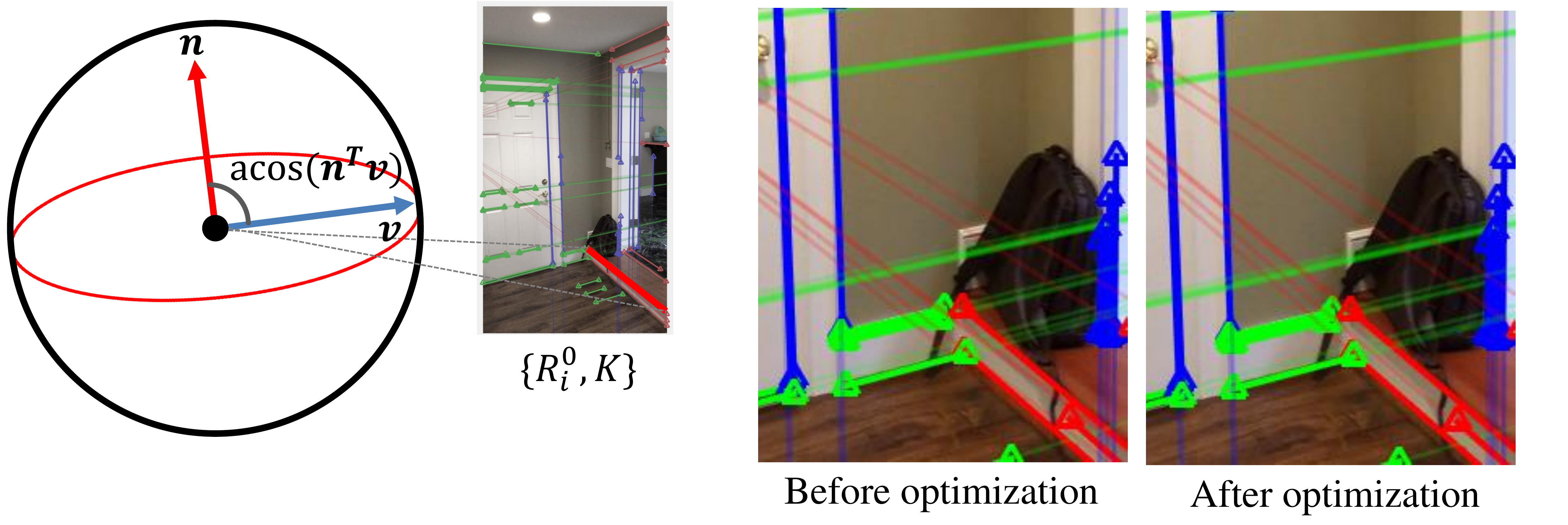}
 \end{center}
 \caption{Rotation refinement. We iteratively update the rotations and
Manhattan line segments until convergence.}  \label{fig:rotation}
\end{figure*}

We follow the VP representation as in~\cite{kroeger2015joint}. The 3D
vanishing directions are represented by the intersections of multiple
{\it interpretation planes}. A interpretation plane is spanned by the
global origin and the two unit vectors that pass both the origin and
endpoints of each line segment in the global space. A homogeneous point
vector $\tilde{\textbf{x}}$ on the $i$-th image is computed by
$R_i^TK^{-1}\tilde{\textbf{x}}$ where $K$ and $R_i$ are the camera
intrinsic matrix and the $i$-th IMU rotation matrix, respectively.

To extract Manhattan frames, we first uniformly discretize the Gaussian
sphere centerd on the optical center of the camera into 10242
directions~\cite{xiao2009image}, and project the interpretation plane of
each line segment to the Gaussian sphere using camera information. Let
$w\in \mathcal{R}^+$ be the length of a line segment on the image. We
accumulate $w$ votes to the discretized bin when the angle difference
between the interpretation plane and the vector on the Gaussian sphere
is less than $0.03$ in radius.  Finally, we normalize the voting map by
the maximum value over all the bins to acquire the normalized Gaussian
sphere.  Figure \ref{fig:frameextraction} shows one example.

The Manhattan frames are extracted by using the votes on the Gaussian
sphere. We use a simple peak extraction algorithm as follow. First, we
find the maximum peak that is near from the gravity direction given by
the IMU sensor ($\textbf{v}_{z}$). Then, we subsample the vectors on the
Gaussian sphere whose angle differences between the peaks and the vector
are less than one degree. From this subset, we find the second peak that
has the maximum votes ($\textbf{v}_x$) and the third vector
($\textbf{v}_y$) that is orthogonal to both $\textbf{v}_x$ and
$\textbf{v}_z$.

We should note that the extracted vanishing directions may contain
errors. The next step improves these vanishing directions as well as the
camera rotation matrices:
	
\section{Rotation refinement}
The global rotation matrices given by the IMU in the consumer
smartphones contain non-negligible errors. Therefore, we refine rotation
matrices using extracted Manhattan frames.

In each image, we declare a line segment as a {\it Manhattan line
segment} if the angle difference between the normal of the
interpretation plane and one Manhattan direction is more than $85$
degrees, and the same angle difference with the other two Manhattan
directions are both less than $85$ degrees. 
The second condition is critical to avoid ambiguous line segments that
correspond to two Manhattan directions.
Given $m$ images and $N(i)$ Manhattan line segments, we refine rotation
matrices by minimizing the following functional:
\begin{eqnarray}
 \sum_{i=1}^{m}{\sum_{j=1}^{N(i)}{{
			\left|\frac{(R_i^TK^{-1}\tilde{\textbf{p}}_{i,j}\times R_i^TK^{-1}\tilde{\textbf{q}}_{i,j})^T}{|R_i^TK^{-1}\tilde{\textbf{p}}_{i,j}\times R_i^TK^{-1}\tilde{\textbf{q}}_{i,j}|}\textbf{v}_{i,j}\right|^2_2}}
	} \nonumber\\
	+ \lambda\sum_{(i,j)\in \mathcal{N}}{\left|R_{i}^TR_{j}-R_{i}^{0T}R_{j}^0u\right|^2_2 }.\label{eq:rotation}
\end{eqnarray}
$K\in \mathcal{R}^{3\times3}$ is the intrinsic camera matrix, and
$R_i\in \mathcal{R}^{3\times3}$ and $R_i^0\in \mathcal{R}^{3\times3}$
are the resulting and initial rotation matrices,
respectively. $\tilde{\textbf{p}}_{i,j}\in \mathcal{R}^{3\times1}$,
$\tilde{\textbf{q}}_{i,j}\in \mathcal{R}^{3\times1}$ and
$\textbf{v}_{i,j}\in \mathcal{R}^{3\times1}$ are homogeneous vectors of
two endpoints and its vanishing direction of $j$-th Manhattan line
segment on the $i$-th image, respectively. $\lambda$ (set by $0.1$ in
our implementation) is a trade-off parameter between the data term and
the smoothness term, respectively.

The first term penalizes the angular difference between the surface
normal direction of the interpretation plane and the assigned vanishing
direction. The second term seeks to make the relative rotation between
nearby frames unchanged during the optimization. The term exploits the
fact that IMU rotation may exhibit long-term accumulation errors, but
are fairly accurate locally.
$\mathcal{N}$ is constructed by the similarity of the $z$-axis of the
camera coordinate frame that is corresponding to the display face
direction of the device. If the angle difference of the $z$-axes is less
than $10$ degrees, two cameras are considered as neighbors.
Note that we repeat the process of extracting Manhattan line segments
and optimizing Eq.\ref{eq:rotation} until convergence (generally three
iterations).  Figure~\ref{fig:rotation} shows that the Manhattan
line segments and vanishing directions are refined over the iterations.

\section{Line tracking}

We match Manhattan line segments between pairs of frames, then find tracks.
When the camera motion is purely panoramic, images are related by the
Homography
$H$~\cite{hartley2003multiple} as
\begin{equation}
\tilde{x_2} = H\tilde{x_1}, \label{eq:homography}
\end{equation}
where $\tilde{x_1}$ and $\tilde{x_2}$ are the point location in the homogeneous coordinate. Since our input is panoramic videos, Eq.~\ref{eq:homography} is roughly satisfied. We find line tracks as follows.

First, we collect all the image pairs whose angle differences of the camera Z-axes
are less than 2 degrees.
For each pair of images, a homography matrix
$H$~\cite{hartley2003multiple} is computed from SURF
matches~\cite{bay2006surf}.

Suppose we seek to match a Manhattan line segment in one frame against
a Manhattan line segment in another.
%
We compute their distance as follows. First, we use the estimated
Homography to warp one line segment to the other frame. Second, for each
end-point of a line segment, compute the distance to a line containing
the other line segment. 4 such distances are computed and we take the minimum as the distance between the two line segments.
%
We declare that a pair of line segments match if 1) this distance is less
than $0.05\min{(h,w)}$; 2) the angle difference is less than 1 degree;
and 3) they are mutually the closest line segment.